\documentclass{article}

\usepackage{arxiv}

\usepackage[utf8]{inputenc} 
\usepackage[T1]{fontenc}    
\usepackage{hyperref}       
\usepackage{url}            
\usepackage{booktabs}       
\usepackage{amsfonts}       
\usepackage{nicefrac}       
\usepackage{microtype}      
\usepackage{bbold}
\usepackage{amsmath}
\usepackage{mathtools}
\usepackage{dsfont}
\usepackage{tabu}
\usepackage[all]{background}

\title{Multi-Objective Autonomous Braking System using Naturalistic Dataset}

\author{
  Rafael Vasquez\\
  Laboratory of Innovations in Transportation\\
  Ryerson University\\
  Toronto, Canada \\
  \texttt{rafael.vasquez@ryerson.ca} \\
   \And
 Bilal Farooq \\
  Laboratory of Innovations in Transportation\\
  Ryerson University\\
  Toronto, Canda \\
  \texttt{bilal.farooq@ryerson.ca} \\
}
\SetBgColor{black}
\SetBgContents{Accepted in the proceedings of IEEE ITSC2019}
\SetBgPosition{-0.5cm,-5.0cm}
\SetBgOpacity{1.0}
\SetBgAngle{90.0}
\SetBgScale{2.0}

\begin{document}
\maketitle

\begin{abstract}
A deep reinforcement learning based multi-objective autonomous braking system is presented. The design of the system is formulated in a continuous action space and seeks to maximize both pedestrian safety and perception as well as passenger comfort. The vehicle agent is trained against a large naturalistic dataset containing pedestrian road-crossing trials in which respondents walked across a road under various traffic conditions within an interactive virtual reality environment. The policy for brake control is learned through computer simulation using two reinforcement learning methods i.e. Proximal Policy Optimization and Deep Deterministic Policy Gradient and the efficiency of each are compared. Results show that the system is able to reduce the negative influence on passenger comfort by half while maintaining safe braking operation.
\end{abstract}

\keywords{Autonomous Braking \and Naturalistic Dataset \and Deep Learning \and Passenger Comfort \and Multi-Objective Optimization}

\section{Introduction}
With the continual advancement of the reinforcement learning paradigm for robotic systems and complex control tasks, the need for large and realistic training datasets is increasing. Autonomous vehicle (AV) simulation platforms provide high-fidelity simulation platforms for data generation and to support the development, training, and validation of autonomous driving systems. One of the focuses of a few of these platforms is the inclusion of pedestrians in order to train agents to react to life-endangering situations that could occur in reality. However, pedestrian road-crossing behaviour is complex and can vary highly in aspects such as waiting time, pose, and path. These intricate behaviours can be difficult to simulate yet carries with it one of the highest priorities when considering pedestrian safety and the inevitable dangerous situations that AVs find themselves in.
        
Another important aspect of transferable autonomous braking systems is having a realistic formulation of the problem during training. This includes accurately representing both the objectives and the mechanism used to reach those objectives. Intuitively, the braking control mechanism of a vehicle resides within a continuous (real-valued) action space. By avoiding the discretization of the action space, we can consider more complex driving behaviours which capture reality more accurately, such as the consideration of passenger comfort.
        
While the pedestrian safety is a top priority, passenger comfort is an important and often overlooked factor in the acceptance and successful implementation of AVs \cite{hartwich_driving_2018}. Naturalness and perceived safety of the AV's driving manoeuvres are two major influences on a rider's comfort. Thus, designing an objective function which accounts for this is crucial for training an acceptable AV braking system.

In our work, we demonstrate the use of highly realistic pedestrian road-crossing data collected by the Virtual Immersive Reality Environment (VIRE) \cite{farooq_virtual_2018} platform to train a multi-objective continuous autonomous braking system in simulation. The proposed system smoothly controls the velocity of the vehicle while reacting safely as necessary in situations where danger is imminent, maximizing passenger comfort while minimizing the chance of an accident occurring. The system is designed by placing the vehicle agent in an uncertain situation in which a simulated pedestrian can cross at any given time depending on the trial data used. The intrinsic randomness of our pedestrian motion data helps to generalize the training. Using the proposed reward function with Proximal Policy Optimization, the agent is able to learn a policy exhibiting highly desirable characteristics. 

The rest of this paper is organized as follows: Section II will include a brief review of pedestrian implementations in popular autonomous training simulation platforms and related work. Section III and IV will contain the details about the design of the proposed autonomous braking system. In Section V, experiment results will be provided along with a discussion of future work in Section VI.

\section{Related Work}
\label{sec:headings}

\subsection{Pedestrians in AV Simulation}
In both research and industry, simulation is used extensively for providing quick and efficient insight regarding concepts, strategies, and algorithms along with their shortcomings before real life testing. Both the safety concerns and the cost of training and testing AV technology are circumvented through the use of simulation. However, publicly available simulation platforms such as AirSim \cite{shah_airsim:_2017} and Gazebo \cite{koenig_design_nodate} do not provide pedestrian simulation, and platforms that do, such as AutonoVi-Sim \cite{best2018autonovi} and CARLA \cite{dosovitskiy_carla:_2017} offer an unrealistic implementation of pedestrian motion and behaviour.
        
In AutonoVi-Sim, pedestrians can be provided destinations within the road network and follow safe traffic rules to navigate to their goal in the default configuration. In order to simulate a jay-walking scenario, pedestrians can be setup to walk into the street in front the agent vehicle. Based on video provided by AutonoVi-Sim, pedestrians walk across the road at a constant velocity and are unable to provide signals regarding intention. In CARLA, pedestrians navigate the streets using a navigation map that conveys a location-based cost. While the cost encourages pedestrians to use sidewalks and road crossings, it allows them to cross a road at any point. While this approach is more stochastic, it should be noted that these implementations emphasize the time and starting point of a crossing occurrence. Equally important aspects of road-crossing behaviour are pedestrian's pose before their cross and their motion during the cross. An autonomous vehicle needs to be able to predict a pedestrian's intention before and during the cross, and neither AutonoVi-Sim or CARLA seem able to provide insight with with regards to this. Is the pedestrian moving at a constant velocity when they cross the road? Where is the pedestrian looking before they cross? Are false-starts (where a pedestrian may begin to cross but change their mind, returning to their starting position) considered? These are a few of the intricacies that pedestrian road-crossing simulation needs to consider, especially in the context of training an autonomous vehicle.

Alternatively, the pedestrian crossing behaviour data collected by VIRE capture these intricacies. By providing respondents with a realistic, immersive and interactive road-crossing experience, their behaviour imitates that of a physical world road crossing.

\subsection{Deep Reinforcement Learning}

Reinforcement learning (RL) is popularly used to train an agent to interact with its surrounding environment. The agent learns by discovering which state-actions yield the most reward through trial-and-error. Deep reinforcement learning (DRL) has recently found success in bigger problems with more state-action pairs by using deep neural networks to approximate the state-action values.

Chae et al. propose an autonomous braking system using DRL \cite{chae_autonomous_2017}. In the study, the agent's sensors receive information about the pedestrian's position and adapts the brake control to the state change in a way that minimizes the chance of a collision. The system is based on a Deep Q-network (DQN). DQN is adapted from the RL method Q-learning \cite{watkins_q-learning_1992} which iteratively searches for an optimal policy by using a Q-value function to calculate the expected sum of rewards for a state-action pair. Q-learning falls short in this case due to the continuous state space, which makes it impossible to find an optimal value of a state-action pair. Instead, DQN approximates the state-action value function using the deep neural network (DNN). This paper formulates the braking scenario as a discrete Markov Decision Process model and defines a discrete action space using a set of four brake control intensities \(\alpha_0, \alpha_{high}, \alpha_{mid}, \alpha_{low}\).

Lillicrap and Hunt noted that DQN can only handle discrete and low-dimensional action spaces and rule out the notion of simply discretizing the action space, as this would cause a overflow of dimensions and actions, making exploration of the action space highly inefficient \cite{lillicrap_continuous_2015}. In their work, they presented a model-free, off-policy actor-critic algorithm using deep function approximators that can learn policies in high-dimensional, continuous action spaces. Based on the deterministic policy gradient (DPG) algorithm  \cite{silver2014deterministic}, the Deep DPG (DDPG) was developed by combining the actor-critic approach found in DPG and insights learned from DQN's recent success. DDPG overcomes the challenge of exploration in continuous action spaces due to its off-policy nature, meaning that the agent decides an action based on another policy. This allows exploration to be treated independently from learning. The algorithm's generality was demonstrated using a racing game which includes the actions acceleration, braking and steering and reported that some agents were able to learn reasonable policies that are able to complete a circuit around the track. More details about DDPG are provided in Section IV.

Zhu et al. explored a reinforcement learning based throttle and brake control approach for a form of the autonomous vehicle following problem \cite{zhu_reinforcement_2017}. Using Neural Dynamic Programming as a kind of model-free reinforcement learning method, a control value corresponding to throttle or brake intensity is derived from a Markov Decision Process framework. It was found that while the agent vehicle could derive near optimal control commands for the autonomous following problem, slight vibrating exists in the control commands during the follow distance regulation process.

Chowdhuri et al. presented a multi-modal multi-task learning approach to deep neural networks. Multi-model learning combines different types of inputs while multi-task learning emphasizes training on side tasks \cite{chowdhuri_multinet:_2017}. They introduced three distinct behaviour modes based on the dataset which are used as a second type of input to the network, allowing for separate driving behaviours to form within a single multi-modal network. These modes are: a) \emph{Direct Mode}, in which the car drives with few obstructions b) \emph{Follow Mode}, in which the car follows a lead car in front of it, and c) \emph{Furtive Mode}, in which the car drives slowly due to perceived boundaries such as bushes alongside the path. This approach inserts behavioural information directory into the network's process stream, allowing for parameter sharing between related modes.

This brief review of related work presents similar attempts to train a neural network to control a vehicle. The emphasis to be made here is that the previous literature either oversimplifies the problem, vehicle control tasks are best described in continuous action spaces, not discrete; or fails to take into consideration an objective for pedestrian perception and passenger comfort. Quirks in a braking (or throttle) control system could be seen by a pedestrian and perceived as unsafe and felt by the passenger and deemed uncomfortable, having negative effect on the acceptance of AVs.
        
\begin{figure}[h]
\centerline{\includegraphics[scale=0.6]{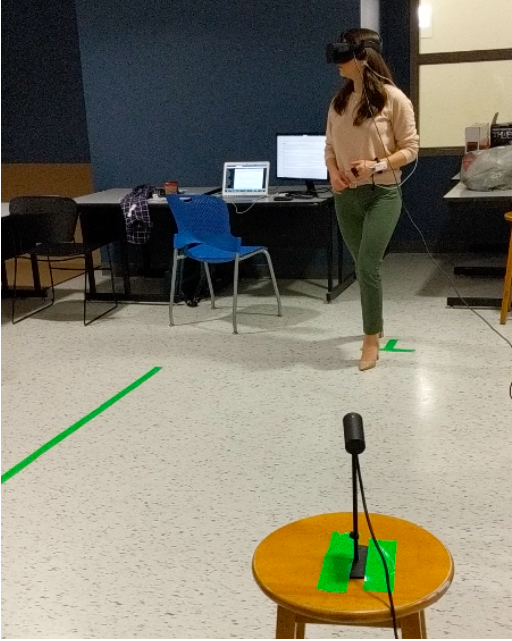}}
\caption{VIRE Pedestrian-AV Physical Environment}
\label{respondent}
\end{figure}

\section{System Description}

\subsection{Data}

\begin{figure}[htbp]
    \centerline{\includegraphics[scale=0.12]{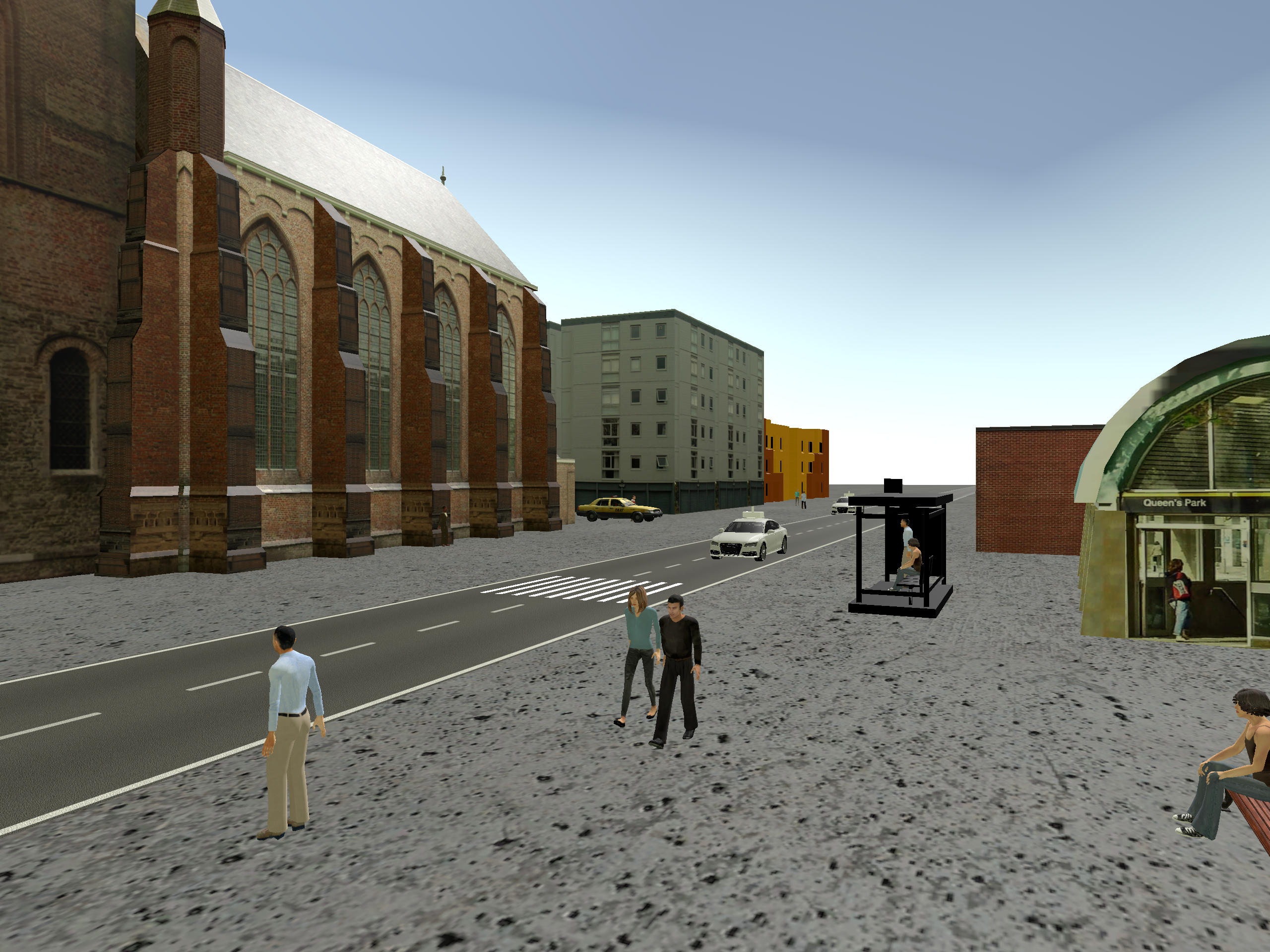}}
    \caption{VIRE Pedestrian-AV Virtual Environment}
    \label{environment}
\end{figure}

Based in the open-source gaming engine Unity, The Laboratory of Innovations in Transportation (LiTrans) has developed VIRE for a range of transportation behaviour experiments that are highly realistic, immersive, interactive and have the capability to collect information about the motion and orientation of the respondents’ interactions with simulated agents (e.g. vehicles, cyclists, etc.). \cite{farooq_virtual_2018} This platform has recently been used to capture pedestrians’ movements in a dense urban environment under varying traffic conditions as well as their general interaction and behaviour towards AVs. \cite{kalatian2019analysis} In the scenarios, respondents stand at the side of a two-lane road with vehicles driving by from one or both directions. They were told to wait for an opportune moment to cross the street without interrupting traffic flow. Running was prohibited because the respondent may hit some physical obstacle in the laboratory or trip themselves with the wires. Fig. \ref{respondent} shows a respondent participating in the experiment, while Fig. \ref{environment} shows the virtual environment the participant is seeing and interacting with through the head-mounted display and sensors. 

From a total of 160 respondents participated in the experiment with a varied demographic which included young teenagers, university students and professionals. The data collected by VIRE for each trial during these experiments include coordinates corresponding to both the position of the respondent and the direction in which they were facing recorded at a time interval of approximately 0.1 seconds. With a total of 2463 individual road-crossing trials, the dataset was found to vary significantly in waiting times, velocities, and completion times. Since each respondent was asked to complete multiple trials, their habits and instincts are captured in the data, giving more naturalism and value to the dataset. We are able to use this data to reproduce the path and pose of the respondent in each trial in simulation using Unity. 

\subsection{Scenario}
One of the biggest challenges for autonomous vehicles is safely handling the risks involved with pedestrians who are considering crossing the road. Acting too cautiously with abrupt and jerky decelerating results in discomfort for the passenger and disruption of traffic flow. Acting aggressively and not decelerating quickly enough can result in an accident. The optimal behaviour results from the ability to recognize the intention of the pedestrian while maintaining safe and comfortable control of the vehicle’s speed. By detecting a pedestrian’s initial movements (e.g. edging towards the sidewalk curb, looking left and right) and considering their relative position, the vehicle should be able to decelerate safely and smoothly.

\begin{figure}[!htbp]
    \centerline{\includegraphics[scale=0.45]{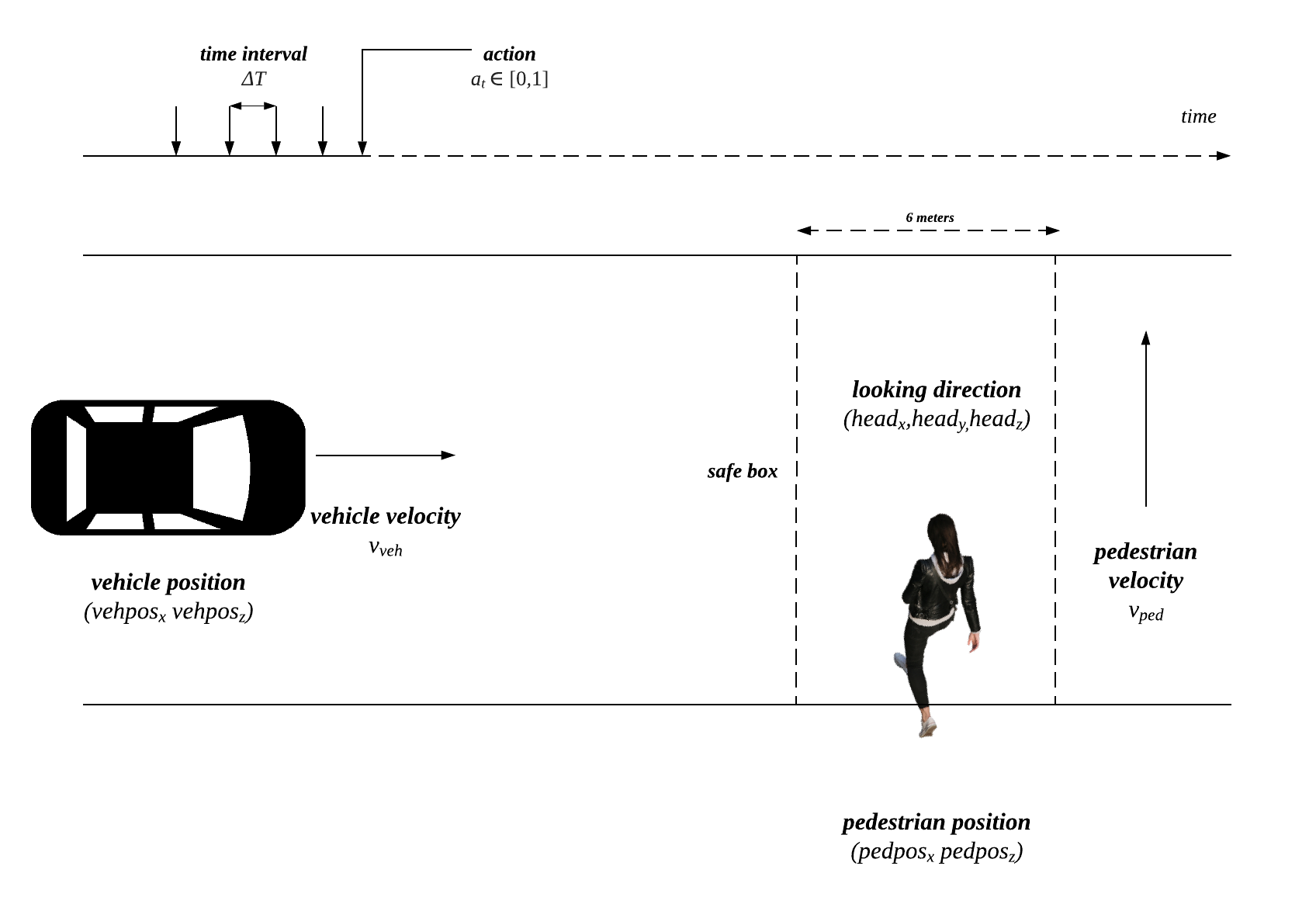}}
    \caption{Illustration of the autonomous braking system}
    \label{diagram}
\end{figure}

In our autonomous braking system, we assume perfect sensor measurements and pedestrian detection. However, stochasticity in the measurements can be included in the future improvements of this work. The vehicle moves at speed \(v_{veh}\) from a position \((vehpos_x, vehpos_y)\) and is provided the relative position to the pedestrian \((pedpos_{x} - vehpos_x, pedpos_y - vehpos_y)\), the looking direction of the pedestrian \((head_x, head_y, head_z)\), and the velocity of the pedestrian \(v_{ped}\), at each time step. Using these observations along with its own velocity, the vehicle decides whether to apply brake, a continuous value between [0, 1]. Fig. \ref{diagram} provides a depiction of the proposed autonomous braking system and scenario.

\section{Deep Reinforcement Learning}

Our autonomous braking system is formalized as a basic reinforcement learning problem in which the agent interacts with an environment in discrete timesteps, receives observations at each timestep \(t\), takes an action \(a_t\) based on policy  \(\pi\) and receives a reward \(r_t\) based on the resulting state \(S_t\). In our case, the possible action that the agent can take is real-valued \(a_t \in \mathbb{R}^N \). In reinforcement learning, and our autonomous braking system, the goal is to learn a policy which maximizes the expected cumulative award that will be received in the future per episode. 

In our system, the vehicle drives along a straight road using the learned policy towards the pedestrian. At the beginning of each episode, the vehicle is driving at an initial velocity \(v_{veh}\) from an initial position \(vehpos_x, vehpos_y\) and the pedestrian is located at an initial position \(pedpos_x, pedpos_y\). A ``safe box'' surrounds the pedestrian's crossing location. If the vehicle is within this boundary while the pedestrian is crossing, it is considered an accident. There are four events that can end an episode:
\begin{itemize}
  \item \textit{Accident}: the vehicle is within the safe box at the time of pedestrian crossing the road
  \item \textit{Pass}: the vehicle passes the pedestrian before the pedestrian has begun to cross
  \item \textit{Cross}: the pedestrian finishes crossing the road
  \item \textit{Stop}: the vehicle comes to a complete stop
\end{itemize}

\subsection{Policy Proximal Optimization}
Proximal Policy Optimization (PPO) \cite{schulman_proximal_2017} is an on-policy algorithm used for environments with either discrete or continuous action spaces. It is a simplification of Trust Region Policy Optimization (TRPO), proposing to use a clipped surrogate loss function to limit the likelihood ratio of old and updated policies \(r_t(\theta) = \frac{\pi_\theta (a_t | s_t)}{\pi_{\theta_old}(a_t|s_t)}\) where \(\pi_\theta\) is a stochastic policy. In TRPO, the surrogate function is maximized subject to a constraint on the size of the policy update. To avoid excessively large policy updates, PPO introduces a penalty to changes to the policy that move the probability ratio away from 1. To include this penalty, the following objective was developed:

\begin{equation*}
    L^{CLIP}(\theta) = \hat{\mathds{E}}_t \Big[\min(r_t(\theta) \hat{A}_t, clip(r_t(\theta), 1-\epsilon, 1+\epsilon)\hat{A}_t)\Big]
\end{equation*}

where epsilon is a small hyperparameter. The second term \(clip(r_t(\theta), 1-\epsilon, 1+\epsilon)\) modifies the surrogate objective in order to remove the incentive for moving \(r_t\) outside of the interval \([1-\epsilon, 1+\epsilon]\). Then, taking the minimum of the clipped and unclipped objective so that the final objective is a lower bound n the unclipped objective. This results in the inclusion of changes in probability ratios only when it makes the objective worse:

\begin{equation*}
    \max_\theta \hat{\mathds{E}}_t \Bigg[\frac{\pi_\theta (a_t | s_t)}{\pi_{\theta_old}(a_t|s_t)} \hat{A}_t \Bigg]
\end{equation*}

\subsection{Deep Deterministic Policy Gradient}
DDPG was developed by combining the actor-critic approach found in DPG and insights learned from DQN's recent success. \cite{lillicrap_continuous_2015} It is is a model-free, off-policy actor-critic algorithm that was proven to learn competitive policies for tasks using low-dimension observations such as cartesian coordinates. It follows a standard reinforcement learning setup as previously described. Unlike PPO, DDPG is inspired by Q-learning, using the greedy policy \(\mu(s) = \arg \max_a Q(s,a)\) to find the optimal action \(a\) for a given state \(s\). Instead of directly implementing Q learning, a modified target network which creates a copy of the actor and critic networks \(Q'(s, a|\theta^{Q'})\) and \(\mu (s|\theta^{\mu'})\) for calculating the target values. The weights of these target networks are then updated, slowly tracking the learned networks which constrains them to change slowly and improves the stability of learning: \(\theta' \leftarrow \tau\theta + (1 - \tau) \theta'\) with \(\tau \ll 1 \). It also implements a finite sized replay buffer \(\mathcal{R}\) which stores transitions sampled from the environment according to the exploration policy in a tuple \((s_t, a_t, r_t, s_{t+1})\). This buffer allows the algorithm to benefit from learning across a set of uncorrelated transitions by sampling a minibatch uniformly at each timestep to update the actor and critic.

\subsection{Reward Function}
As the most important component of our system, the reward function determines the behaviour of the brake control. We must formulate it such that it accurately describes the characteristics the system's behaviour should ultimately exhibit. Given the multi-objective nature of our system, our task in defining the reward function becomes more difficult; we are attempting to optimize two possibly conflicting factors, avoiding dangerous situations with pedestrians and providing the passenger with a comfortable experience. To this end, we propose a reward function which captures both safe and comfortable braking behaviour:

\begin{gather*} 
 r_t = 
-(\eta v_{veh_t})\textbf{1}(S_t = accident) \\
- (\beta v_{veh_t})\\
\displaystyle - (\mu a_t) \Bigg | { \frac{d^2 v_{veh}}{dt^2}} \Bigg |\\
\eta, \beta, \mu > 0
\end{gather*}
  
where at time step \(t\), \(v_t\) is the velocity of the vehicle, \(dist_t\) is the distance in meters between the vehicle and the pedestrian, and \(a_t\) is the value of the brake action chosen and \textbf{1}\((x=y)\) has a value of 1 if the statement inside is true and 0 otherwise. The first term is included from the reward function proposed by Chae et al. \cite{chae_autonomous_2017}, describing the penalty given to the agent should in the case of an accident. The penalty is proportional to the velocity of the vehicle, reflecting the severity of the accident and therefore encouraging the vehicle to slow down in the case that an accident is unavoidable. The second and third terms are our own contributions. The second encourages the vehicle to use caution and prevents the agent from abusing the stochastic nature of our dataset and unintentionally learning to take the risk of driving passed the pedestrian before they begin to cross the road. The final term is a function of the jerk of the vehicle at time step \(t\). Mathematically, jerk is the second derivative of velocity with respect to time. Physically, it is felt as the increasing or decreasing force on the body \cite{eager_beyond_2016}. The absence of jerk has been found to be an important characteristic of comfortable driving \cite{bellem_comfort_2018, bellem_objective_2016} and thus, our reward function aims to minimize this force, penalizing the vehicle as a function of the jerk and the intensity of the brake action. The constants \(\eta\), \(\beta\), and \(\mu\) are weight parameters to control the trade-off between each objective. 

\section{Experiments}

\subsection{Simulation Setup}
For simulations, we used the game engine platform Unity along with its open-source plugin ML-Agents Toolkit \cite{juliani_unity:_2018}  to train the PPO agent. We designed an environment in which our pedestrian dataset that was also collected using Unity, is easily transferable. In each episode, the pedestrian is situated at the same initial position, but the pedestrian crossing trial used is chosen at random from a subset of the dataset at random each episode. The rest of the dataset is used to test the model. Trials vary in waiting times, crossing velocities and head movements, helping to train our system for generality. The vehicle is initially located 160 meters away from the pedestrian, and has an initial velocity of \(v_{init} = 11.11m/s\) (40 \(km/h)\). The pedestrian safe box size is 3 meters wide, covering any situation in which the vehicle is within 3 meters of the pedestrian. 

\subsection{Training}
The neural network implementation for PPO consists of a fully-connected feed-forward network with three hidden layers and the hyperparameter configuration shown in Table \ref{hyperp}.

\begin{table}[h]
\parbox{.45\linewidth}{
\centering
\caption{PPO hyperparameter configuration}
\label{hyperp}
\begin{tabular}{| c |c|} 

\hline
Batch size & 64 \\ 
\hline
Buffer size & 10240 \\ 
\hline
\# of Hidden Units & 256 \\ 
\hline
Time Horizon & 1024 \\ 
\hline
Learning Rate & \(10^{-3}\) \\ 
\hline
Gamma & 0.99\\ 
\hline
\(\eta\) & 0.1 \\ 
\hline
\(\beta\) & 0.01 \\ 
\hline
\(\mu\) & 0.01 \\ 
\hline
\end{tabular}
}
\hfill
\parbox{.45\linewidth}{
\centering
\caption{DDPG hyperparamter configuration}
\label{ddpg}
\begin{tabular}{| c | c |} 
\hline
Replay Buffer Size & 10240 \\ 
\hline
Minibatch Size & 128 \\ 
\hline
Gamma & 0.99 \\ 
\hline
Tau & \(10^{-3}\) \\ 
\hline
Actor Learning Rate & \(10^{-4}\) \\ 
\hline
Critic Learning Rate & \(10^{-4}\) \\ 
\hline
\(\eta\) & 0.1 \\ 
\hline
\(\beta\) & 0.01 \\ 
\hline
\(\mu\) & 0.01 \\ 
\hline
\end{tabular}
}
\end{table}

Our implementation of DDPG includes two hidden, fully connected layers with 256 and 128 nodes for the actor and critic networks respectively. Its configuration is shown in Table \ref{ddpg}.

A total of three models were trained: two of the models include the passenger comfort objective, one trained using PPO and the other with DDPG to compare their learning. A third model was trained with PPO, but did not include the passenger comfort objective in the reward function in order to contrast their action trajectories. These models are outlined in the Table \ref{algoconf}.

\begin{figure}[htbp]
\centerline{\includegraphics[scale=0.9]{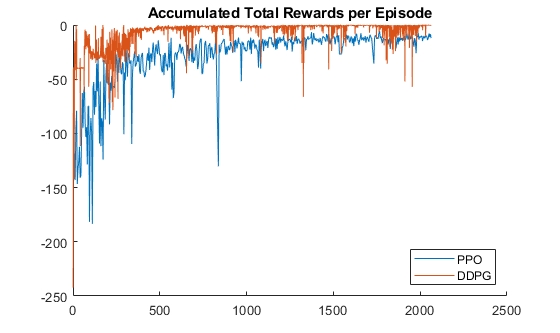}}
\caption{Cumulative Environment Reward}
\label{cumulative}
\end{figure}

\begin{table}[!ht]
\centering
\caption{Algorithm configurations}
\label{algoconf}
\begin{tabular}{| c | c |} 
\hline
 Algorithm & Passenger Comfort \\ [0.2ex] 
\hline\hline
PPO\(_1\) & Included \\ 
\hline
PPO\(_2\) & Not Included \\ 
\hline
DDPG & Included \\ 
\hline
\end{tabular}
\end{table}
During training, a total of 2320 pedestrian-crossing trials were utilized. The data was split into a training and testing set. The training set contains a random combination of 1856 trials while the testing set contains a random combination of the remaining 464 trials. Fig. \ref{cumulative} shows a plot of the total accumulated rewards spanning every episode during training using PPO\(_1\) and DDPG. It can be observed that DDPG not only converges much earlier than PPO\(_1\), but also consistently attains high total reward thereafter. 

\subsection{Results}
\begin{figure}[!h]
\centerline{\includegraphics[scale=0.85]{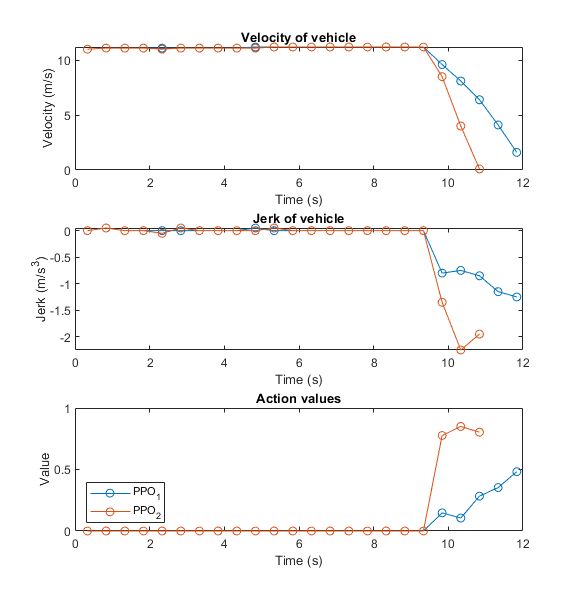}}
\caption{Profile of velocity, jerk and actions in an example episode against the same road-crossing trial}
\label{jerk}
\end{figure}

During evaluation, each model produced reported an absence of any accidents with the pedestrian. This was to be expected as the vehicle was given ample time to come to a complete stop in all cases. In our case, it is more important to note the braking profiles produced by the agents trained with PPO\(_1\) and PPO\(_2\). Fig. \ref{jerk} shows a detailed trajectory of brake actions for each agent for one example case.  Fig. \ref{jerk} (a) shows the velocity of the vehicle over time, as it approaches the pedestrian. The jerk of the vehicles are presented in (b) while (c) shows the brake action values applied. Both agents begin their deceleration at the same time, but execute actuation differently. It is clear that the agent trained using PPO\(_2\) decelerated steeply, using higher brake action values to do so and in doing so, produced relatively severe jerk. Alternatively, the agent trained using PPO\(_1\) has a curved decent in velocity as a result of its use of weaker brake action values applied at the beginning part of its deceleration and strengthening as the vehicle nears the pedestrian. Most importantly, the mean jerk force produced is halved, implying a much more comfortable experience for those in the vehicle. This example shows that while both agents exhibit safe braking operation by avoiding a collision, it is clear which one provides a better travelling experience. The inclusion of vehicle jerk in the reward function results in more comfortable ride for passengers while maintaining reasonable safe braking operation.

\section{Conclusions and Future Work}
We have demonstrated the use of a naturalistic dataset to train and test a multi-objective autonomous braking system while establishing the significance of an objective which maximizes vehicle passenger comfort. We have also shown that the system is capable of reducing the negative influence on passenger comfort (i.e. jerk force) by half while maintaining a safe braking policy. There are many directions we can take with this work. The presented work can be built upon to configure more complex pedestrian situations such as incorporating multiple pedestrians along the road. Along with the introduction of a throttle action for the vehicle, passenger comfort and pedestrian safety can be further evaluated. Different weather conditions and their adverse effects on the vehicle such as loss of traction can also be introduced in the physics-based Unity environment. Additionally, the dataset can be used toward modeling pedestrian path and pose as well as developing and evaluating pedestrian intention prediction techniques.

\newpage
\bibliographystyle{unsrt}  
\bibliography{ITSC_abbre}  

\end{document}